\documentclass[journal]{IEEEtran}
\usepackage{amsmath,amssymb,amsfonts}
\usepackage{algorithm}
\usepackage{algorithmic}
\usepackage{pifont,bbding,multirow}
\usepackage{arydshln,booktabs}
\usepackage{array}
\usepackage{hyperref}
\usepackage{graphicx}
\usepackage{cite}
\usepackage[caption=false,font=normalsize,labelfont=sf,textfont=sf]{subfig}
\usepackage{textcomp}
\usepackage{stfloats}
\usepackage{url}
\usepackage{verbatim}
\usepackage{xcolor}

\usepackage{multirow}
\usepackage{booktabs,soul}
\usepackage{amsmath,bbding,pifont}

\definecolor{mypink1}{rgb}{0.858, 0.188, 0.478}
\definecolor{green}{rgb}{0.158, 0.8, 0.3}

\title{Expertized Caption Auto-Enhancement for Video-Text Retrieval}

\author{Baoyao Yang ~\IEEEmembership{Member,~IEEE}, Junxiang Chen\IEEEauthorrefmark{1}, Wanyun Li, Wenbin Yao and Yang Zhou
\thanks{B. Y. Yang is with the School of Computers, Guangdong University of Technology, P. R. China;}
\thanks{J. X. Chen, W. Y. Li, W. B. Yao, and Y. Zhou are with WeChat, Tencent, P. R. China;}
\thanks{$^*$Corresponding author: J. X. Chen (\textit{E-mail:caryjxchen@tencent.com})
}
\thanks{\textit{This work has been submitted to the lEEE for possible publication. Copyright may betransferred without notice, after which this version may no longer be accessible.}}
}

\begin{document}
\maketitle

\begin{abstract}
Video-text retrieval has been stuck in the information mismatch caused by personalized and inadequate textual descriptions of videos.
The substantial information gap between the two modalities hinders an effective cross-modal representation alignment, resulting in ambiguous retrieval results. Although text rewriting methods have been proposed to broaden text expressions, the modality gap remains significant, as the text representation space is hardly expanded with insufficient semantic enrichment.
Instead, this paper turns to enhancing visual presentation, bridging video expression closer to textual representation via caption generation and thereby facilitating video-text matching.
While multimodal large language models (mLLM) have shown a powerful capability to convert video content into text, carefully crafted prompts are essential to ensure the reasonableness and completeness of the generated captions. 
Therefore, this paper proposes an automatic caption enhancement method that improves expression quality and mitigates empiricism in augmented captions through self-learning.
Additionally, an expertized caption selection mechanism is designed and introduced to customize augmented captions for each video, further exploring the utilization potential of caption augmentation.
Our method is entirely data-driven, which not only dispenses with heavy data collection and computation workload but also improves self-adaptability by circumventing lexicon dependence and introducing personalized matching. The superiority of our method is validated by state-of-the-art results on various benchmarks, specifically achieving Top-1 recall accuracy of \textbf{68.5\%} on MSR-VTT,  \textbf{68.1\%} on MSVD, and \textbf{62.0\%} on DiDeMo. Our code is publicly available at \textcolor{blue}{\textit{{https://github.com/CaryXiang/ECA4VTR}}}.
\end{abstract}


\section{Introduction}
\label{sec:intro}
\begin{figure}[t]
\centering
\includegraphics[width=0.98\columnwidth]{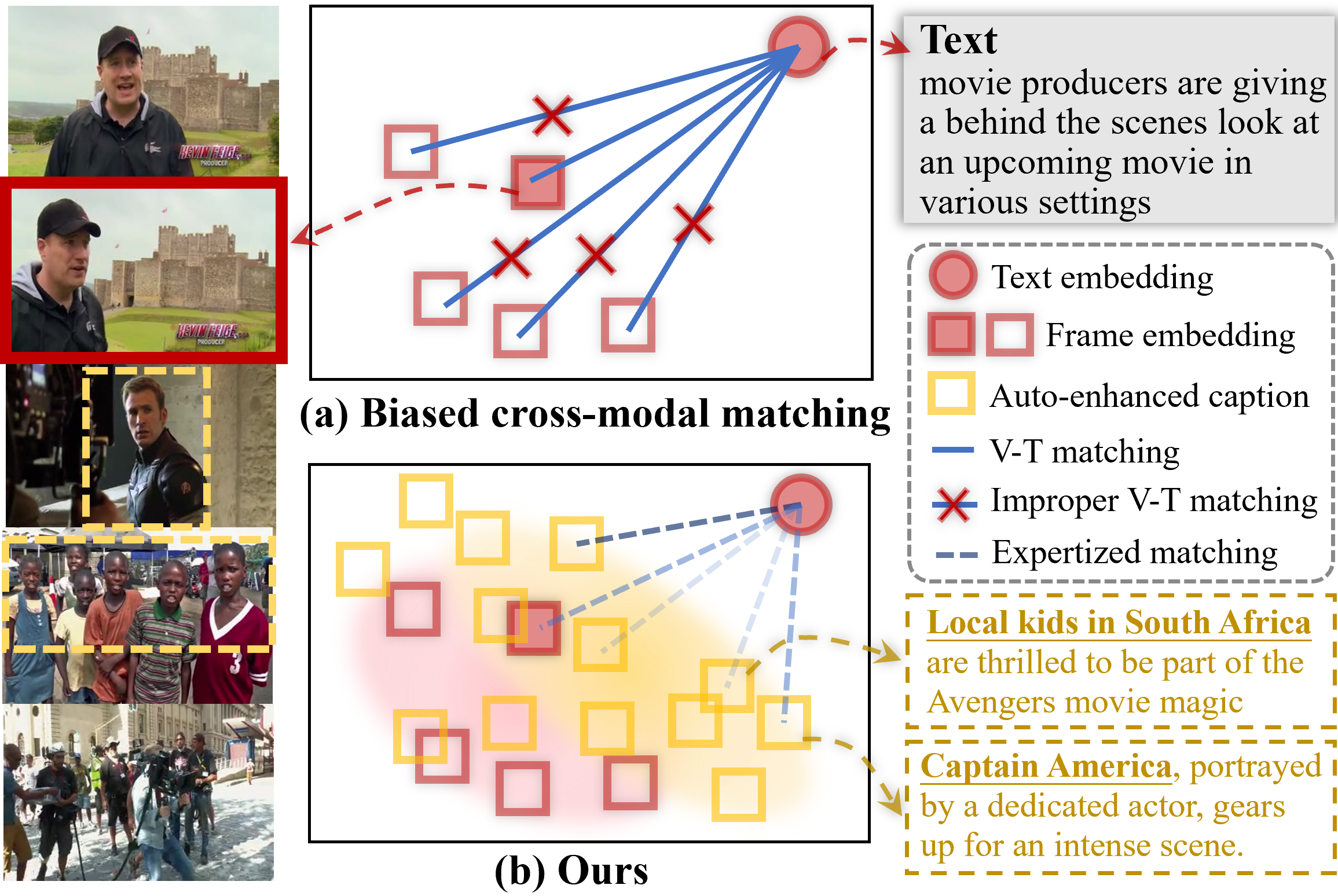} 
\caption{Comprehension gap between videos and texts}
\label{fig:gap}
\end{figure}

Video-text understanding has become a new mainstream research with the population of short videos on social networks. In virtue of the learning ability of large models, multi-modal pre-training models \cite{li2023unmasked,chen2024vast} are developed and achieve initial success in video-text retrieval (VTR) tasks. However, a stunning amount of data and computing resources are the essential basis to support the effectiveness of these pre-training models.
A lighter and more feasible approach to achieve VTR is adapting the image-text pretrained model to video-text scenarios \cite{luo2022clip4clip,xue2022clip}. In this approach, 
the visual embeddings are adapted to the video field by introducing temporal information \cite{sun2022long-form,Chen2022LiteVLEV} or refining the visual-text matching strategies \cite{gao2021clip2tv,xue2022clip}. 
% They adapt the visual embeddings to the video field by incorporating temporal information \cite{sun2022long-form,Chen2022LiteVLEV} or refining the visual-text matching strategies \cite{gao2021clip2tv,xue2022clip}. 
But unlike image-text pairs, the textual descriptions of videos tend to be overly simplistic and one-sided, describing partial \textit{(even a single)} frames of the videos. 
The improper alignment between one-sided descriptions and multifaceted video embeddings would confuse text understanding \textit{(Fig. \ref{fig:gap}~(a))}, resulting in biased cross-modal retrieval, and a bottleneck in performance improvement arises.

% The improper alignment between texts and 
% These one-sided descriptions would enlarge the modality gap, confusing cross-modal assignment

% mproperly matched to multifaceted video embeddings \textit{(Figure \ref{fig:gap} (a))}, resulting in biased cross-modal retrieval, and a bottleneck in performance improvement arises.

% As shown in Figure \ref{fig:gap}, the text describing a specific frame \textit{(red box)} is 

Frame sampling is one proper solution to handle the improper alignment issue. For instance, ClipBERT \cite{lei2021less} sparsely samples video frames to meet the text descriptions. X-CLIP~\cite{ma2022x} further cuts text descriptions into words, searching for video frames for fine-grained cross-modal matching. As video information is sparsely divided and broken, these methods are limited in comprehensively utilizing and exploring deep relations across videos and texts. 
Another branch to promote video-text understanding is data augmentation. Extra data, such as HD-VILA-100M \cite{xue2022advancing} and LF-VILA~\cite{sun2022long-form}, are collected to expand the receptive field for model adaptation, thereby enriching matching knowledge of the two modalities.
To address the substantial data collection burden, subsequent research has explored text rewriting strategies to enhance expressions on the text side. For example, Wu \textit{et al.} \cite{wu2023bidirectional} rewrite texts using lexical analysis, in which attributes are extracted to enhance expressive ability in the text modality. She \textit{et al.}~\cite{yu2025she} further ameliorate text presentation using syntax-tree, emphasizing key words to assist better VTR. 
% Large language models (LLMs), such as ChatGPT and LLaMA \cite{touvron2023llama}, are additionally introduced for the text rewriting tasks~\cite{liu2023visual,fan2024improving}. 
Besides generating textual descriptions, text enhancement is executed in the feature space~\cite{croitoru2021teachtext,wang2024text}, providing embedding variants for cross-modal alignment. 
Despite achieving a retrieval precision increase of up to 3\% with the aid of supplementary textual expressions, minimal additional semantics are introduced during the text-to-text rephrasing processes, which hinders further reduction of the modality gap. This restricts significant breakthroughs in cross-modal retrieval capability.

% Although up to a 3\% increase in retrieval precision is achieved with the assistance of supplementary texts, few additional semantics are introduced during these rephrasing processes, preventing further reduction of the modality gap; consequently, cross-modal retrieval capability has limited break through.

Instead, this paper suggests enhancing the representations on the video side, deriving captions from videos to meet the expression modes and semantic information of texts. The rich semantics of videos are preserved, and video-text matching would be simplified due to the reduced modality gap achieved by this directional data augmentation.
% This approach facilitates a more straightforward video-text matching process through directional data augmentation.
Although multimodal Large Language Models (mLLMs) have shown a decent ability in video understanding \cite{cap4video++}, generating appropriate video captions for VTR remains challenging, as it is revealed that the performance of mLLMs is query-sensitive. The writing quality of mLLMs relies heavily on lexicon engineering~\cite{parashar2024neglected} and hand-crafted prompts \cite{momeni2023verbs}. Recent studies~\cite{bansal2024videocon,wang2024havtr} have been mired in the meticulous work of designing better prompts for caption augmentation, limited in human empiricism.

\begin{figure}[t]
\centering
\includegraphics[width=0.95\columnwidth]{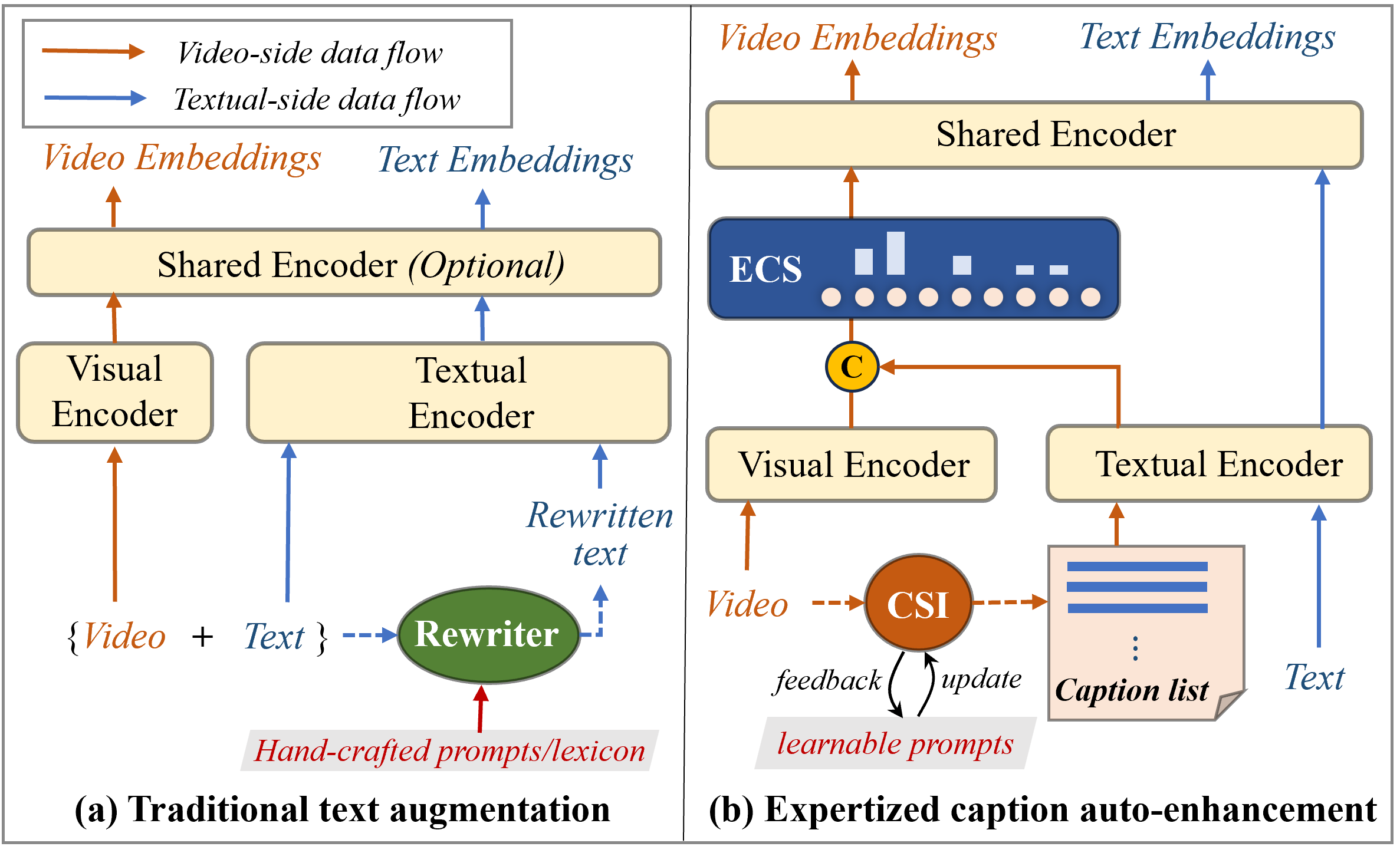} 
\vspace{-0.5em}
\caption{Comparison of our method to the traditional}
\label{fig:overview}
\end{figure}

Inspired by automatic prompt optimization (APO) \cite{pryzant2023automatic,yang2023dynamic}, this paper proposes an Expertized Caption Auto-Enhancement (ExCae) method, as shown in Fig. \ref{fig:overview} (b).
% ExCae adaptively generates and selectively personalizes captions for VTR, thereby alleviating the additional burden of data acquisition and eliminating the complexities associated with manual prompt design.
ExCae incorporates two main components: Caption Self-Improvement (CSI) and Expertized Caption Selection (ECS) modules, which adaptively generates and selectively personalizes captions for VTR. The two modules provide ExCae with the advantage of alleviating the additional burden of data acquisition and eliminating the complexities associated with manual prompt design.
%respectively search for the best generation and utilization of video captions. 
Specifically, in the CSI module, a \textit{Captioner} implemented by a multimodal LLM (mLLM) is asked to derive captions from videos. We optimize the query prompt of the \textit{Captioner} to ensure the generated captions contain rich content and diverse perspectives. To overcome the interpretability loss of prompt in existing gradient-based APO, this paper introduces a \textit{Prompt Engineer}, which gradually improves the query prompts of the \textit{Captioner} via repetitively inquest and output adjustment.
% , which \textit{\textbf{adaptively generates}} and \textit{\textbf{personalized select}} captions for video-text retrieval, \underline{\textit{releasing extra work of data acquisition}} and \underline{\textit{avoiding the trouble of manual prompt design.}}
% Specifically, a Caption Self-improvement (CSI) module is designed to derive captions with rich content and diverse perspectives from videos. A \textit{Prompt engineer} is introduced in the CSI module, which gradually adjusts and improves the query prompts via self-learning. 
With the enriched expressions on the video side \textit{(video+video-derived captions)}, the ECS module consisting of multiple learnable experts is designed to automatically specify appropriate expressions for cross-modal matching. 
Seeing Fig. \ref{fig:gap} (b), video representation space dilates with caption augmentation, and the cross-modal information gap is reduced after expertized selection. This indicates that the proposed method improves the generalization ability through caption augmentation while enhancing the cross-modal retrieval ability with adaptive screening. As a fully data-driven process, sampling bias and empiricism that may confuse the cross-modal matching are avoided. We further encapsulate ExCae as a \underline{\textit{plug-in unit}}, which successfully improves the retrieval performance of current methods, validating ExCae's flexibility and adaptability.

\noindent The contributions of this work are summarized as follows:
\begin{itemize}
\item We propose reducing the modality comprehension gap for VTR by enhancing visual presentations via self-improved caption derivation from videos. Video captions are progressively refined, bridging video expression closer to text space while eliminating the grunt workload of data acquisition.
% by bringing video expression closer to textual representation via
% eliminating the grunt workload of data acquisition and prompt design in caption augmentation through self-improvement and expertized selection, thereby mitigating empirical bias in supplemental texts while facilitating cross-modal matching without extra data.

\item We design an automatic video-text learning framework for VTR, in which video expressions are expanded using a \textit{Prompt Engineer} and are matched with texts under the guidance of learnable experts. The \textit{Prompt Engineer} facilitates effective and multi-angle caption generation through a self-improvement strategy that mitigates empirical bias in prompt design, while the learnable experts personalize the appropriate expression angles for cross-modal matching. 

\item Experimental results show that our method outperforms \textit{(or is comparable to)} current works even without using extra data, especially surpassing the best benchmark on the MSR-VTT by \textbf{3.8\%}. This paper also taps the application potentials of ExCae in video-text retrieval via extensive analytical and plug-in experiments.
\end{itemize}

% We propose eliminating the grunt workload of data acquisition and prompt design in caption augmentation through self-improvement and expertized selection, thereby mitigating empirical bias in supplemental texts while facilitating cross-modal matching without extra data.

\section{Related Work}
\label{sec:related_work}

Developing and pre-training video foundation models (VFMs) \cite{li2023unmasked,wang2022internvideo} enhances the comprehension of video data, demonstrating efficacy in downstream tasks such as video-text retrieval. However, the effectiveness of VFMs heavily relies on huge amounts of data and computing resources. 
% CLIP \cite{radford2021learning} is an exceptionally powerful cross-modal model that has achieved significant success in image-text retrieval. 
Much of the current work in video-text retrieval turns to a lighter modeling approach, adapting image-text retrieval models, such as CLIP \cite{radford2021learning}, to video-text tasks.
For example, CLIP4Clip \cite{luo2022clip4clip} and CLIP2TV \cite{gao2021clip2tv} 
transform CLIP's embeddings into effective video representations through various frame aggregation strategies. 
In addition, X-CLIP \cite{ma2022x} proposes using an independent encoder for each frame to derive video representations.
Video embeddings are further enhanced in TS2-Net \cite{liu2022ts2}, which dynamically selects video tokens with richer information. 
Transfer learning is another appropriate solution to adapt knowledge from image to video field, based on which Jin \textit{et al.} \cite{jin2024mv} introduce a transformation block to adaptively transfer knowledge in the pre-trained CLIP from image-text to video-text.
This adaptation is also achieved by a teacher-student framework in TeachClip \cite{holistic}.
Considering the feature redundancy issue, DRL \cite{wang2022disentangled} designs a loss in channel decorrelation regularization to improve the quality of video representation. 
However, these methods suffer from ill-suited cross-modal matching due to the information imbalance between videos and texts. 
% utilize the image knowledge encapsulated in CLIP, transforming it into effective video representations through different image aggregation networks.
% In addition, TS2-Net \cite{liu2022ts2} dynamically adjusts the video token sequence, selecting tokens with richer information to enhance video representation. 

% X-CLIP \cite{ma2022x} employs an independent encoder for each frame, using a network that aggregates frames multiple times to generate the final video representation. 

% DRL \cite{wang2022disentangled} introduces an additional loss to address the feature redundant issue. 

To improve cross-modal matching, existing works, such as BIKE \cite{wu2023bidirectional} and SHE-NET \cite{yu2025she}, enhance text expression with lexical and syntax analysis. Although Large language models (LLMs), such as ChatGPT and LLaMA \cite{touvron2023llama}, are additionally introduced in the text rewriting tasks~\cite{du2024reversed,liu2023visual,fan2024improving}, visual details remain deficient in the rewritten texts through these text-to-text literal restatements.
Generating captions from videos as auxiliaries is a brighter approach for text enhancement. These methods form new pairs using videos and derived captions to promote cross-modal representation learning \cite{cap4video++}. 
However, carefully designed prompts are necessary, and later works manually craft various query prompts, asking LLMs to describe empirically selective concepts, such as special objectives \cite{parashar2024neglected} or actions \cite{momeni2023verbs}. 
The hand-drafted lexicon and prompts severely limit the rewrite quality, motivating us to explore a self-learning caption enhancement method. 
To the best of our knowledge, this paper is a \textit{\textbf{pioneer}} in investigating \underline{\textit{\textbf{automatic caption enhancement}}} for videos to facilitate the comprehension of videos and texts.

\begin{figure}[t]
\centering
\includegraphics[width=0.98\columnwidth]{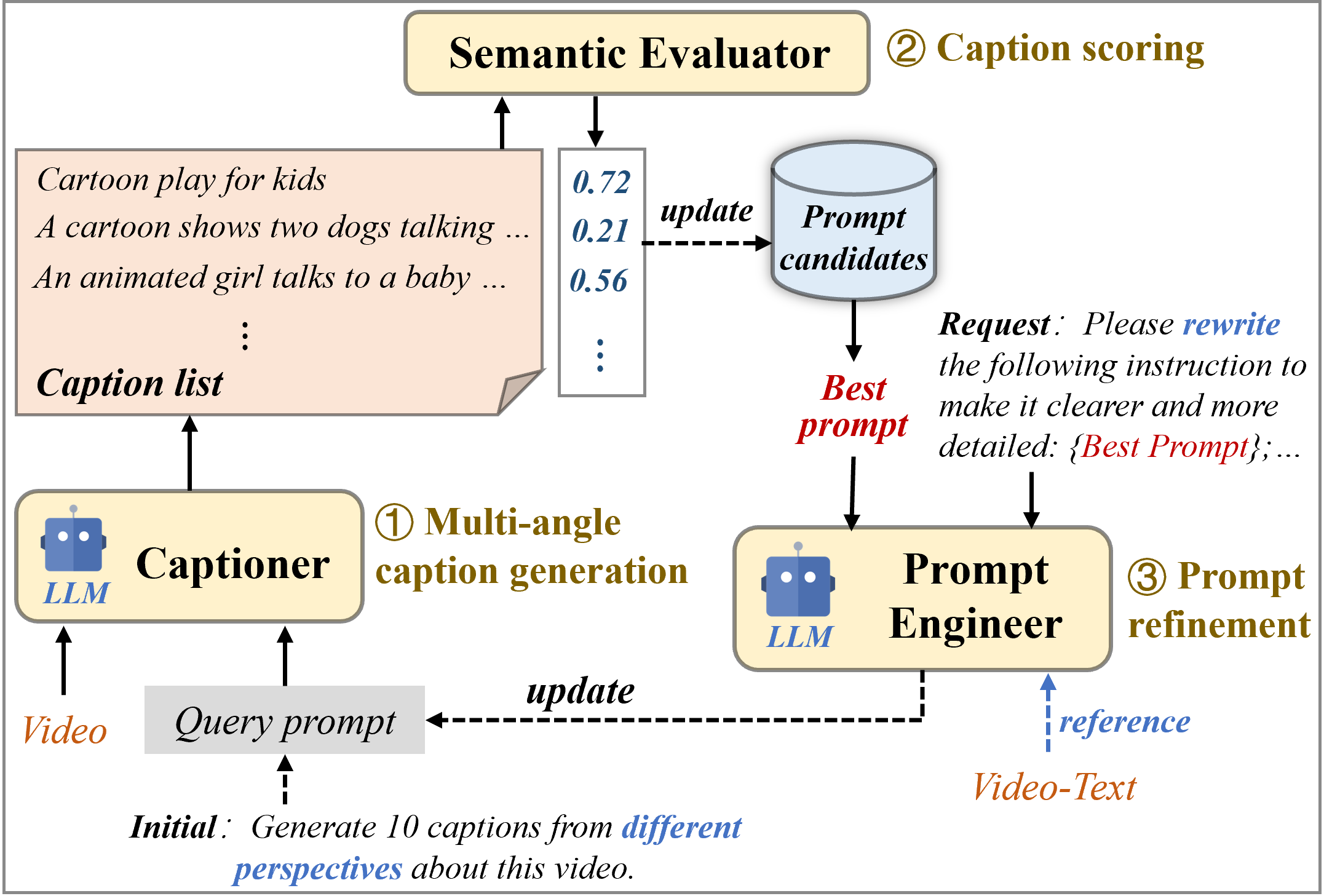} 
 \vspace{-0.5em}
\caption{Illustration of the CSI module}
\label{fig:CSI}
\end{figure}

\section{Method}
Unlike traditional text augmentation that rewrites captions on the text side, our method (ExCae) automatically derives captions from videos by the CSI module and adaptively selects appropriate expressions (ECS module) to encode video for video-text matching \textit{(see Fig. \ref{fig:overview})}. Video and text embeddings would be aligned with standard cross-modal contrastive loss, as most existing SOTA methods \cite{liu2022ts2,luo2022clip4clip}. CSI and ECS are self-learning modules whose details are introduced as follows. 

\subsection{Caption Self-improvement (CSI)}
In the CSI module, a \textit{Captioner} is included, which is an mLLM asked to generate a set of captions based on a query prompt to enhance video understanding. As intermediaries for video-text matching, the captions derived from videos are expected to be informative and text-oriented. To achieve this goal, CSI performs an iterative prompt refinement (Fig.~\ref{fig:CSI}), gradually updating the query prompts of the \textit{Captioner} by repeatedly inquiring a \textit{Prompt Engineer}. 
% In each iteration, the current query prompt is evaluated from two aspects (caption scoring and semantic assessment), resulting in the update of the prompt candidates' pool.
In each iteration, the current query prompt is evaluated by caption scoring, an approximate assessment of VTR performance, resulting in the update of the prompt candidate.
The \textit{Prompt engineer} then rewrites the current best prompt, whose output is used as the latest query prompt of the \textit{Captioner}. This prompt update iteration would end when no new candidates are added. Pseudo code is provided in Algorithm \ref{alg:CSI}. 

% The main flow of CSI is illustrated in Figure~\ref{fig:CSI}.
% Initially, the 

% The main flow is illustrated in Figure~\ref{fig:CSI}, which is an iteration process gradually refining the query prompt. Specifically, a \textit{Captioner} is asked to generate multiple captions to describe the video from \textit{\textbf{different perspectives}}. We then evaluate the semantics of the generated captions, scoring them to update the prompt candidates' pool. A \textit{Prompt engineer} is introduced to rewrite the current best prompt, whose output is used as the latest query prompt of the \textit{Captioner}. 

% This prompt update iteration would end when no new candidates are added to the prompt candidates' pool. Pseudo code is provided in Algorithm \ref{alg:CSI}.

\paragraph{\textbf{\textit{Caption scoring}}} 
We hope that the adjustment of the query prompt will be positively correlated with the increase in VTR results throughout the iterative process. Directly evaluating VTR results as feedback for prompt refinement involves multimodal encoder optimization, where slight fluctuations in embeddings can significantly affect the direction of prompt optimization. 
Therefore, this paper designs a scoring mechanism that evaluates the quality of generated captions, implicitly reflecting the performance of VTR.
Denote the query prompt for the \textit{Captioner} at iteration $t$ as $p^t$. The generated caption list for video $v_i$, using $p^t$ is denoted as $\mathbf{c}_i^t = \{c_{i,1}^t, c_{i,2}^t, \ldots, c_{i,K}^t\}$, where $K$ is the number of captions per video.
To balance semantic fidelity and diversity, $\mathbf{c}_i^t$ is evaluated from two perspectives:    
\begin{enumerate}
    \item \textbf{Semantic Consistency}: Ensures $\mathbf{c}_i^t$ semantically align with the original text $t_i$, mitigating the presence of cluttered and vacuous descriptions.
    \item \textbf{Caption Diversity}: Encourages multi-angle descriptions in $\mathbf{c}_i^t$, avoiding redundancy to maintain the original intention of expression enhancement.
\end{enumerate}

\noindent The overall score for iteration $t$ is computed as:
\begin{equation}
Score^t = \mathbb{E}_{i,k}[sim(\psi(c^t_{i,k}),\psi(t_i))] + div(\psi(\mathbf{c}^t_i))
\label{eq:score}
\end{equation}
where $k \in [1,K]$ is the caption index, and $\psi(\cdot)$ represents a semantic extractor. $sim(\cdot)$ calculates the cosine similarity on semantics. $div(\cdot)$ quantifies diversity within $\mathbf{c}^t_i)$:
\begin{equation}
\hspace{-0.5em} div(\psi(\mathbf{c}^t_i)) = \frac{1}{K(K-1)} \sum_{p \neq q} \left( 1 - sim \left( \psi(c_{i,p}^t), \psi(c_{i,q}^t) \right) \right)
\label{eq:div}
\end{equation}
% are similarity and diversity calculations. In specific, $sim(\cdot)$ is computed by consine similarity, \textit{i.e.}, $sim(\psi(c^t_{i,k}),\psi(t_i))=\frac{\psi(c^t_{i,k}) \cdot \psi(t_i)}{\|\psi(c^t_{i,k})\| \|\psi(t_i)\|}$.

% , implemented by \textit{M3-embedding}~\cite{chen2024bge}.

% where $\psi(\cdot)$ is a semantic embedding extractor (e.g., \textit{M3-embedding}~\cite{chen2024bge} ), $\text{sim}(\cdot)$ measures cosine similarity, and $\text{div}(\cdot)$ quantifies diversity.

\noindent The higher value of $Score^t$ indicates better alignment with the original text and greater caption diversity. 
If $Score^t$ surpasses the historical records, the prompt candidate is updated by $p^t$, and the best prompt used for prompt refinement is updated accordingly.

\paragraph{\textbf{\textit{Prompt refinement}}}
Current prompt engineering normally encodes the query prompts and adjusts the prompt tokens/embeddings through gradient descent \cite{pryzant2023automatic,yang2023dynamic}. Although this parametric fine-tuning strategy is convenient for optimization, the adjusted embeddings could hardly be precisely re-projected to the text space, losing linguistic meaning and interpretability. Consequently, their generalization in video-text-related applications is limited.
Inspired by language rewrites \cite{fan2024improving}, this paper introduces a \textit{Prompt engineer} to rewrite the query prompt. Take the current best prompt $p^t$ as input; the \textit{Prompt engineer} is asked to rewrite $p^t$ with reference to the corresponding video-text pairs $(x,t)$ involved in this iteration. The rewritten prompt then inquires the \textit{Captioner} in the next iteration, as shown in Fig. \ref{fig:CSI}.

\begin{algorithm}[t]
\caption{Caption Self-improvement (CSI) algorithm}
\begin{algorithmic}[1]
\STATE \textbf{Input:} Video-text pairs \((x,t)\), Initial query prompt \(p_0\), Maximum iteration $T$
\STATE \textbf{Output:} \(Best\_prompt\) 

% \STATE Put $p_0$ in Prompt candidate' pool; 
\STATE \textbf{Initialization}:
\STATE ~~~~\(Best\_prompt \gets p_0;\) 
\STATE ~~~~\(Best\_score \gets -\infty;\) 
% \STATE Generate the caption list \(\mathbf{c}\) for videos using $p_0$;
% \STATE Calculate \(Score^0\) using Equation \eqref{eq:score}
% \WHILE{\(Score^t-Score^t\)}
\FOR{$t = 1, 2, \dots, T$}
    % \STATE \# Prompt refinement:
    \STATE $p_t \gets PE(Best\_prompt)$; ~~~~~// Refine prompt
    \IF{\(Best\_prompt == p_t\)} 
    \STATE break; ~~~~~~~~~~~~~~~~~~~~~~~~~~// No update
    \ENDIF
    % \STATE \# Generate caption lists for videos: 
    \STATE $\mathbf{c} \gets Captioner(x,p_t)$; ~~~~~~~~// Generate Captions
    \STATE Calculate \(Score^t\) using Equation \eqref{eq:score};
    \IF{\(Score^t > Best\_score\)}
    \STATE \(Best\_score \gets Score^t;\)
    \STATE \(Best\_prompt \gets p_t;\) ; 
    \ENDIF
    % \STATE Refining using \textit{Prompt Engineer}: \textit{Captioner($Best\_prompt$)}; 
    % \STATE \textbf{Step 1: Generate captions}
    % \STATE \(C_t \gets Captioner(p_t)\)
    
%     \STATE \textbf{Step 2: Evaluate captions and update prompt candidates}
%     \STATE \(caption\_scores, semantic\_assessment \gets f_1(C_i), f_2(C_i)\)
%     \STATE Update prompt candidates: \(P\_candidates \gets (caption\_scores, semantic\_assessment)\)
    % \STATE Prompt refinement: \textit{PromptEngineer($Best\_prompt$)};
%     \STATE \(best\_prompt \gets select\_best\_prompt(P\_candidates)\)
%     \STATE \(refined\_prompt \gets E.refine\_prompt(best\_prompt, V, C_i)\)
    
%     \STATE \textbf{Check if prompt has changed}
%     \IF{\(refined\_prompt = P_i\)}
%         \STATE break
%     \ELSE
%         \STATE \(P_i \gets refined\_prompt\)
%     \ENDIF
\ENDFOR
\end{algorithmic}
\label{alg:CSI}
\end{algorithm}

\subsection{Expertized Caption Selection (ECS)}
With the optimized CSI module, video-text pairs are updated as $((v,\tilde{c}),t)$, where $(v,\tilde{c})$ represents the video data $v$ and its generated captions $\tilde{c}$ using the optimal prompt, and $t$ denotes the original text corresponding to $v$.
As displayed in Fig. \ref{fig:overview} (b), $(v,\tilde{c})$ and $t$ are the video-side and text-side input data, respectively. 
A straightforward approach for cross-modal representation learning is directly matching the encoded representations of $(v,\tilde{c})$ and $t$.
However, the information dilation of video modality brings confounded knowledge while improving generalization, increasing the difficulty of cross-modal matching \textit{(see Fig. \ref{fig:gap})}.

Considering that the derived captions describe video content from multiple perspectives, we propose inserting an ECS module designed based on a mixture of experts to facilitate cross-modal matching. Write the outputs of visual encoders as $e_v=[e_{v1},e_{v2},...]$, which consists of $N+K$ elements, where $N$ is the number of sampled frames from videos, and $K$ is the number of derived captions. The ECS module is composed of multiple expert networks with learnable parameters ($[f_1, f_2, ...,f_M]$), each of which refers to a description angle. $M$ denotes the total number of experts. A router $r_m$ is allocated to each expert $f_m$ as a gate, which is automatically learned to explore the most matched visual expression fusion for each video-text pair. Through this individualized video expression screening and fusion, the expression that is overly different from the corresponding texts is filtered on the video side, thereby narrowing the video-text modality gap.
% In the ECS module, each expert represents one aspect of video expressions, and a router is automatically learned to explore the most matched expression fusion for each video-text pair. 
% The ECS module is composed of multiple expert networks with learnable parameters ($[f_1, f_2, ...,f_M]$), each of which refers to a description angle. $M$ denotes the total number of experts. A router $r_m$ is allocated to each expert $f_m$ as a gate to automatically select visual expressions meeting this expert. 
%Write the outputs of visual and textual encoders as $e_v$ and $e_t$, respectively. 
Mathematically, the visual codes are residually updated by a mixture of experts' outputs:
\begin{equation}
\tilde{e}_v = \frac{1}{M}\sum_m f_m ( r_m(e_v) \otimes e_v) + e_v
\end{equation}
where $r_m(e_v)$ outputs attention weights of $(N+K)$ video-side expressions for each sample. Specifically, a top-$R$ filter is applied on $r_m(e_v)$, which activates partial expressions to learn a specific expert. $\otimes$ denotes the dot-product operation.  
As visual and textual embeddings will be aligned after further encoding by a shared encoder, parameters of the ECS module ($f_m$ and $r_m$) are optimized along with other encoding networks under the supervision of textual information.

\begin{table*}[th]
    \centering
    \caption{Comparison with SOTA methods on MSR-VTT dataset \textit{(underline marks the best ViT-based results)}}
      \vspace{-0.8em}
    % \small
  %  \begin{tabular}{lp{1.4em}<{\centering}p{2em}<{\centering}|p{2.2em}<{\centering}p{2.2em}<{\centering}cc|p{2.2em}<{\centering}p{2.2em}<{\centering}cc}
    \begin{tabular}{llc|cccc|cccc}
        \toprule
        \multirow{2}{*}{Method} & \multirow{2}{*}{Backbone} &  Extra &  \multicolumn{4}{c|}{text-to-video}   & \multicolumn{4}{c}{video-to-text}   \\
        \cline{4-11}
          & &  data  & R@1$\uparrow$ &  R@5$\uparrow$ &  R@10$\uparrow$ & MeanR$\downarrow$ & R@1$\uparrow$ &  R@5$\uparrow$ &  R@10$\uparrow$    & MeanR$\downarrow$   \\
        \midrule
       \multicolumn{1}{l}{ \textit{\textbf{Pre-trained foundation model}} } \\
         InternVideo \cite{wang2022internvideo}  & \textit{ViT-H/14} & \ding{51} & 55.2 &	79.6	& 87.5	&10.7 & 57.9	& 79.2	& 86.4	& 	7.5\\
         UMT-L  \cite{li2023unmasked}  & \textit{ViT-L/14} & \ding{51} &  58.8 &  81.0 & 87.1 & -- & 58.6 & 81.6 & 86.5 & --\\

         CLIP-VIP \cite{xue2022clip} & \textit{ViT-B/16} & \ding{51}  &  57.7	    & 80.5	   & 88.2	& --  & -- & --  & -- & --\\
         
          mPLUG-2 \cite{xu2023mplug} &  \textit{ViT-L/14} & \ding{51} & 53.1 & 77.6 & 84.7  & -- & --  & -- & --  & -- \\
          VAST \cite{chen2024vast} & \textit{ViT-G/14}  & \ding{51} & 63.9 & 84.3 & 89.6 & --  & -- & --  & -- & --\\
          
        \midrule
        \textit{\textbf{ViT-based}} \\
         CLIP2TV \cite{gao2021clip2tv} & \textit{ViT-B/16} & \ding{51}& 49.3& 74.7 &83.6 &13.5& 46.9 &75.0 &85.1 &10.0 \\
        DRL \cite{wang2022disentangled}  & \textit{ViT-B/16} & \ding{51}   & 49.4& 76.4 &84.2 & 13.2& 47.0 &77.1 &84.4 &9.2 \\
        TS2-Net \cite{liu2022ts2} & \textit{ViT-B/16} & \ding{55}  &47.8 &76.8 &85.2 & 13.7& 47.8 &76.0& 84.6 &8.5 \\
        Clip4Clip \cite{luo2022clip4clip}  & \textit{ViT-B/16} & \ding{51} & 46.4  & 72.1 &  82.0 &  14.7  & 45.4 &  73.4  & 82.4 & 10.7 \\
        X-CLIP \cite{ma2022x}  & \textit{ViT-B/16} &  \ding{55}  & 49.3&   75.8 &  84.8 &   12.2 &  48.9 &  76.8&   84.5 &   8.1\\
        DMAE \cite{jiang2023dual}  & \textit{ViT-B/16} &  \ding{55}   &  49.9 & 75.8 & 85.5 &  12.5  & 49.6 & 76.3& 85.0&  8.5  \\
        Cap4Video++ \cite{cap4video++}  & \textit{ViT-B/16}  &  \ding{55} & 52.3&  76.8&  85.8 & 11.5 & 50.0 & 75.9 & 86.0 & 7.8 \\
        TeachClip \cite{holistic}  & \textit{ViT-B/16} &  \ding{55} & 48.0 & 75.9 & 83.5 & --  & -- & --  & -- & --\\
    % \hline
    \cdashline{1-11}
    \multirow{4}{*}{ExCae \textit{(\textbf{ours})}} &  {\textit{ViT-B/16}} & \ding{55} & 55.0 & 84.6 & 91.3 & 6.0 & 56.1 & 84.2 & 90.9 & 7.4 \\
       & {\textit{ViT-L/14}} & \ding{55}   & 60.5	 & 87.3 & 	95.0	 & 3.8	 & 62.5	 & 89.0	 & 94.1 & 	4.1 \\
      & {\textit{ViT-H/14}}  & \ding{55}  & 62.2	 &88.2 &	93.7 &	3.9	& 65.1	&89.4&	94.0	&3.8  \\
       & {\textit{ViT-G/14}} & \ding{55}   &  \underline{\textbf{67.7}} & 	 \underline{\textbf{92.7}}	 &  \underline{\textbf{96.2}}	 &  \underline{\textbf{2.9}}	 &  \underline{\textbf{69.3}}	 &  \underline{\textbf{92.5}}	 &  \underline{\textbf{96.3}}	 &  \underline{\textbf{2.3}}   \\
         \bottomrule
    \end{tabular}
    \label{tab:msr_comparison}
\end{table*}

\section{Experiment}
\subsection{Datasets}
\noindent\textbf{\textit{MSR-VTT}} \cite{xu2016msr} is a large-scale dataset containing 10,000 video clips. Each video is described in 20 English sentences. This paper uses
9,000 clips for training and 1,000 clips for testing, as in the setup of previous work \cite{Liu2019UseWY}.
% , each of which is annotated with 20 English sentences. Around 29K unique words are contained in all sentences. Following previous work \cite{Liu2019UseWY}, we use 9,000 clips for training and 1,000 clips for testing.

\noindent\textbf{\textit{MSVD}} \cite{chen2011collecting} consists of 1,970 videos. Each video contains around 40 English captions. In total, around 80K captions are provided. The dataset is derived by 1,200/100/670 for training, validation and testing, respectively. 

% The samples are split by 1,200/100/670 for training, validation and testing, respectively. 

\noindent\textbf{\textit{DiDeMo}} \cite{anne2017localizing} contains over 10,000 videos downloaded from Flickr.
Each video has a caption recording detailed information on camera movement, temporal transition indicators, and activities. 
% collected from Flickr. 
% contains over 10,000 videos collected from Flickr. Each video is described by natural language, with detailed information on camera movement, temporal transition indicators, and activities. 
We split the dataset into training, validation and test sets, each containing 8,395, 1,065 and 1,004 videos, respectively.

\begin{table*}[th]
    \centering
      \caption{Comparison with SOTA methods on MSVD and DiDeMo datasets \textit{(underline marks the best ViT-based results)}}
     \vspace{-0.8em}
       %  \begin{tabular}{lp{1.4em}<{\centering}p{2em}<{\centering}|p{2.2em}<{\centering}p{2.2em}<{\centering}cc|p{2.2em}<{\centering}p{2.2em}<{\centering}cc}
    \begin{tabular}{llc|cccc|cccc}
        \toprule
        \multirow{2}{*}{Method} & \multirow{2}{*}{Backbone} &  Extra &  \multicolumn{4}{c|}{text-to-video}   & \multicolumn{4}{c}{video-to-text}   \\
        \cline{4-11}
        &  & data  & R@1$\uparrow$ &  R@5$\uparrow$ &  R@10$\uparrow$ & MeanR$\downarrow$ & R@1$\uparrow$ &  R@5$\uparrow$ &  R@10$\uparrow$    & MeanR$\downarrow$  \\
        \midrule
        \multicolumn{10}{c}{\textbf{MSVD dataset}} \\
        \midrule
       \multicolumn{1}{l}{ \textit{\textbf{Pre-trained foundation model}} } \\
         InternVideo \cite{wang2022internvideo}  &  \textit{ViT-H/14} & \ding{51}  & 58.4 &	\textbf{84.5} &	90.4	& 8.2 &  76.3	& \textbf{96.8} &	98.7	& 2.1\\
         UMT-L  \cite{li2023unmasked}  &  \textit{ViT-L/14} & \ding{51} &58.2 & 83.9 & 89.6 & -- &  --  & -- & --  & -- \\
     %   VAST \cite{chen2024vast}  & \ding{51} \\
         
        \midrule
        \textit{\textbf{ViT-based}} \\
         CLIP2TV \cite{gao2021clip2tv} & \textit{ViT-B/16}   & \ding{51} & 50.2 & 79.8 & 87.9 & 8.6 & --& --& --& --\\
         DRL \cite{wang2022disentangled}  & \textit{ViT-B/16} & \ding{51} & 50.0 & 81.5 &89.5 &--&68.7 &92.5 &95.6&--\\
        Clip4Clip \cite{luo2022clip4clip} & \textit{ViT-B/16} & \ding{51} &  47.2&  77.7  & --&9.1   & 63.2 & 87.2&-- & 4.2   \\
        X-CLIP \cite{ma2022x}  & \textit{ViT-B/16} &  \ding{55} & 50.4&  80.6 & -- &  8.4&  66.8 & 90.4 &  -- &  4.2 \\
        % Cap4Video \cite{wu2023cap4video} &  \textit{ViT-B/16} &  \ding{55}&  51.8 &80.8& 88.3&8.3 & 70.0&	93.2	&96.2	&	2.4\\
        Cap4Video++ \cite{cap4video++} &  \textit{ViT-B/16} &  \ding{55}&  51.8 &80.8& 88.3&8.3 & 70.0&	93.2	&96.2	&	2.4\\
     %\hline
       \cdashline{1-11}
     \multirow{4}{*}{ExCae \textit{(\textbf{ours})}} &  {\textit{ViT-B/16}}  & \ding{55}   & 51.3  & 78.9	& 86.5	 & 10.6 & 60.6	 & 83.9	 &	88.7	& 5.5 \\
       & {\textit{ViT-L/14}} & \ding{55}   & 53.6&	80.3&	86.9	&10.4&	67.6&	90.9&	96.5	&3.1 \\
      & {\textit{ViT-H/14}} & \ding{55}&  56.5 &	82.2	&88.4	&9.4	&74.1	&93.4&	97.2	&2.3  \\
      &  {\textit{ViT-G/14}} & \ding{55} & \underline{\textbf{59.2}}		& \underline{83.4}		& \underline{\textbf{90.7}}		& \underline{\textbf{5.1}}	&	\underline{\textbf{76.9}}	&	\underline{96.6}	&	\underline{\textbf{99.0}}		& \underline{\textbf{1.6}} \\
        \midrule
         \multicolumn{10}{c}{\textbf{DiDeMo dataset}} \\
        \midrule
        \multicolumn{1}{l}{ \textit{\textbf{Pre-trained foundation model}} }\\
         InternVideo \cite{wang2022internvideo}  &  \textit{ViT-H/14} & \ding{51}  & 57.9	& 82.4	& 88.9	&	9.2 & 59.1 & 	81.8 &	89.0&7.2\\
         UMT-L  \cite{li2023unmasked}  & \textit{ViT-L/14} & \ding{51} & 70.4 & 90.1 & \textbf{93.5} & -- & \textbf{65.7}  & \textbf{89.6} & \textbf{93.3} & --\\
         CLIP-VIP \cite{xue2022clip} & \textit{ViT-B/16} & \ding{51} &  55.3 & 82.0	  & 89.3	& --  & -- & --  & -- & --\\
         mPLUG-2 \cite{xu2023mplug} & \textit{ViT-L/14} & \ding{51} & 56.4 & 79.1 & 85.2  & -- & --  & -- & --  & -- \\
        VAST \cite{chen2024vast} &  \textit{ViT-G/14} & \ding{51} & \textbf{72.0} & \textbf{89.0} & 91.4 & -- & --  & -- & --  & --\\
             
        \midrule
        \textit{\textbf{ViT-based}} \\
       DRL \cite{wang2022disentangled}  & \textit{ViT-B/16} & \ding{51}  &  49.0 & 76.5 & 84.5 & 11.5& 49.9 & 75.4 & 83.3 & 7.9 \\
        Clip4Clip \cite{luo2022clip4clip} & \textit{ViT-B/16}  & \ding{51} & 44.8  &73.4 & -- & 13.5  &44.7 & 74.0  &-- &10.6\\
        X-CLIP \cite{ma2022x}  & \textit{ViT-B/16} &  \ding{55} & 47.8 &79.3& --& 12.6& 47.8& 76.8 & --&10.5\\
        Cap4Video++ \cite{cap4video++}& \textit{ViT-B/16} &  \ding{55}  & 52.5&  	80.0&  	87.0	&  	10.3	&  51.2	&78.5&  	87.4	&7.3	\\      
        TeachClip \cite{holistic} & \textit{ViT-B/16} &  \ding{55}  & 43.7 & 72.7 &  & -- & -- &-- & -- & -- \\
        \cdashline{1-11}
      \multirow{4}{*}{ExCae \textit{(\textbf{ours})}} &  {\textit{ViT-B/16}} & \ding{55} &  53.9  & 80.6 &	87.3  &	 	9.9 & 54.0 & 81.1 & 87.5 &  6.9  \\
       & {\textit{ViT-L/14}} & \ding{55} & 56.9 &	80.9 &	87.9 &	6.5	 & 58.8	 & 83.2	 & 90.0 & 4.8 \\
       &  {\textit{ViT-H/14}} & \ding{55}  &  60.1	 &83.3	&90.0	 &5.9 &	61.0	 &85.6	 &92.3	 &4.5 \\
        &  {\textit{ViT-G/14}} & \ding{55}  & \underline{62.0}	&\underline{85.5}	& \underline{92.3} & \underline{\textbf{4.9}}	& \underline{63.7} &	\underline{87.9}	& \underline{\textbf{93.3}} &	\underline{\textbf{4.1}} \\
      \bottomrule
    \end{tabular}
  
    \label{tab:msvd_comparison}
\end{table*}

\subsection{Implementation Details}
\label{sec:implement}

\paragraph{\textbf{Preprocessing}}
Following the standard text preprocessing \cite{radford2021learning}, we split the textual data, including original texts and video-derived captions, into word tokens with a max length of 70 using a CLIP tokenizer. For videos, we sample 8 frames from each item as visual data.

\paragraph{\textbf{Model Setup}}
This paper reports the results of our method on multiple backbones, \textit{i.e.}, ViT-B/16, ViT-L/14, ViT-H/14, ViT-G/14, compared to SOTA video-text retrieval methods, showing the effect of model scale on performance as well. Ten captions are derived for each video. 
Sixteen experts are learned in the ECS module, two of which are activated to select appropriate representations in both training and inference processes.  In the training phase, the CSI module is pre-learned to acquire the optimal prompt for caption generation, while the ECS module is co-trained alongside the backbone parameters.
In the evaluation phase, multiple captions are derived from each video using the optimal prompt, together with the video itself as video-side expressions.
% \textit{(See \underline{Supplemental Materials} for the learned best prompt and examples of video-derived captions.)}
The \textit{Captioner} and the \textit{Prompt engineer} are implemented by GPT-4o, a large cross-modal inference model recently released by OpenAI in 2024. The initial prompt asking the \textit{Captioner} is set as ``\textit{Generate 10 captions from different perspectives about this video.}" The semantic embedding extractor is Equation \eqref{eq:score} is  implemented by \textit{M3-embedding}~\cite{chen2024bge}.
In addition to the comparison experiment, we encapsulate ExCae as a plug-in unit and test its effect when introduced into existing video-text retrieval methods; specifically, we enhance video expression via CSI and professionalize the video-text matching process by ECS.
Ablation studies and analytical experiments are conducted on MSR-VTT using ViT-G/14 as the base model.

\paragraph{\textbf{Environment and Evaluation Metrics}}
Our model is trained on 8 NVIDIA  A800 GPUs in a batch size of 16. The pretrained weights from CLIP \cite{radford2021learning} are used to initialize the text and video encoder in ExCae. The model is trained using the Adamax optimizer with a learning rate of 4e-6. 
Experimental results are evaluated by standard video-text retrieval metrics: Recall at Rank K (R@K) and Mean Rank (MeanR). R@K denotes the percentage of correct samples in the top K retrieved samples. In this paper, K is set to 1, 5 and 10, respectively, the same as in previous works~\cite{xue2022clip,jiang2023dual}.
MeanR records the average rank of correct items in the retrieved list. The higher value in R@K means better retrieval performance. In contrast, a lower MeanR is preferred in VTR.
% and the lower va

\subsection{Comparision to State-of-the-arts}
The results of our method are compared to those of the SOTAs, including pre-trained foundation models like InternVideo \cite{wang2022internvideo}, UMT-L \cite{li2023unmasked}, mPLUG-2 \cite{xu2023mplug}, VAST \cite{chen2024vast}, and ViT-based methods that adapt the image-text models to the video-text task. Among the ViT-based methods, CLIP2TV \cite{gao2021clip2tv}, DRL \cite{wang2022disentangled}, Clip4Clip \cite{luo2022clip4clip} and CLIP-VIP \cite{xue2022clip} require the acquisition of auxiliary data for post-pretraining and/or adaptation. In contrast, TS-Net \cite{liu2022ts2}, X-CLIP \cite{ma2022x}, DMAE \cite{jiang2023dual}, Cap4Video++ \cite{cap4video++} and TeachClip \cite{holistic} are learned without extra data.

Results in Table \ref{tab:msr_comparison} show that, on the MSR-VTT dataset, our method is superior to all existing methods, including pre-training foundation models and the ViT-based methods. 
% relies on any s
% only to the ViT-based methods but also to the pre-training models relying on large-scale pre-training data. 
The results of existing methods show that the Top-1 recall accuracy could hardly transcend 50\% without assistance from extra data. Instead, great improvement \textit{(up to 10\%)} is obtained when additional data is involved, either they are used for pre-training or fine-tuning. 
% Our outstanding performance also does not request any extra data.
Despite not using any extra data, the Top-1 recall accuracy of our method (ExCae) boosts to \textbf{55.0}\% and \textbf{56.1}\% in text-to-video and video-to-text retrieval, respectively, when using ViT-B/16 as the base model.
We are also encouraged by the observation that ExCae's performance demonstrates an increase in correlation with the enlarged scale of the backbone model, resulting in an around \textbf{7\%} rise of R@1 from ViT-L/14 to Vit-G/14. Our results have comprehensively surpassed the current SOTA pre-training models, such as mPLUG-2 and VAST, when using the Vit-G/14 as the backbone. The rising trend of performance indicates the potential for further improvement of our method through scaling up the base model.
Similar results are obtained in experiments on MSVD and DiDeMo datasets, as recorded in Table~\ref{tab:msvd_comparison}. Our results are comparable to or even superior to those of pre-training models. 
Compared to existing ViT-based methods, ExCae shows a resounding victory, with Top-1 recall accuracy suppressing the best one by \textbf{7\%} on MSVD and \textbf{13\%} on DiDeMo. 
These results validate that ExCae is robust to VRT tasks for different data sources. 

\begin{figure*}[t]
	\centering
	\includegraphics[width=0.95\textwidth]{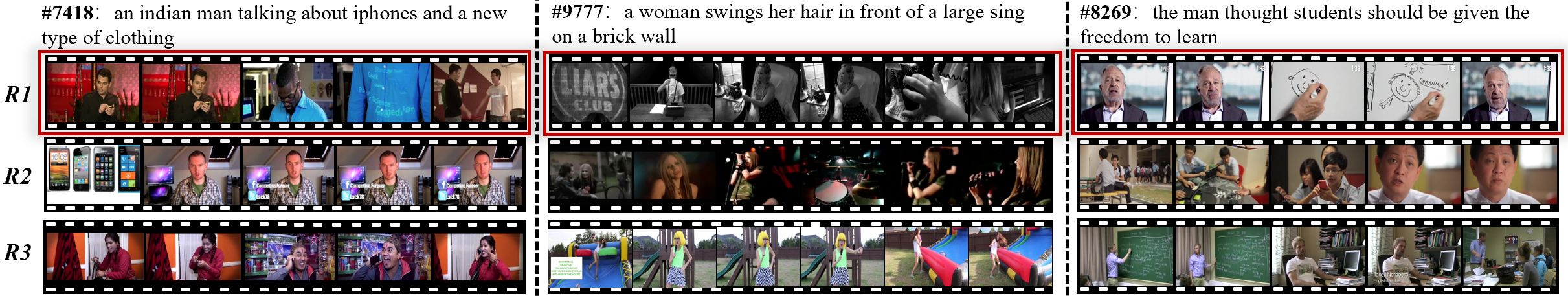}
        \vspace{-0.5em}
	\caption{Examples of text-to-video retrieval results on MSR-VTT dataset.  \textit{(The ground truths are marked in a red box.)}}
	\label{fig.comp_t2v_example}
\end{figure*}

\begin{figure*}[t]
	\centering
	\includegraphics[width=0.95\textwidth]{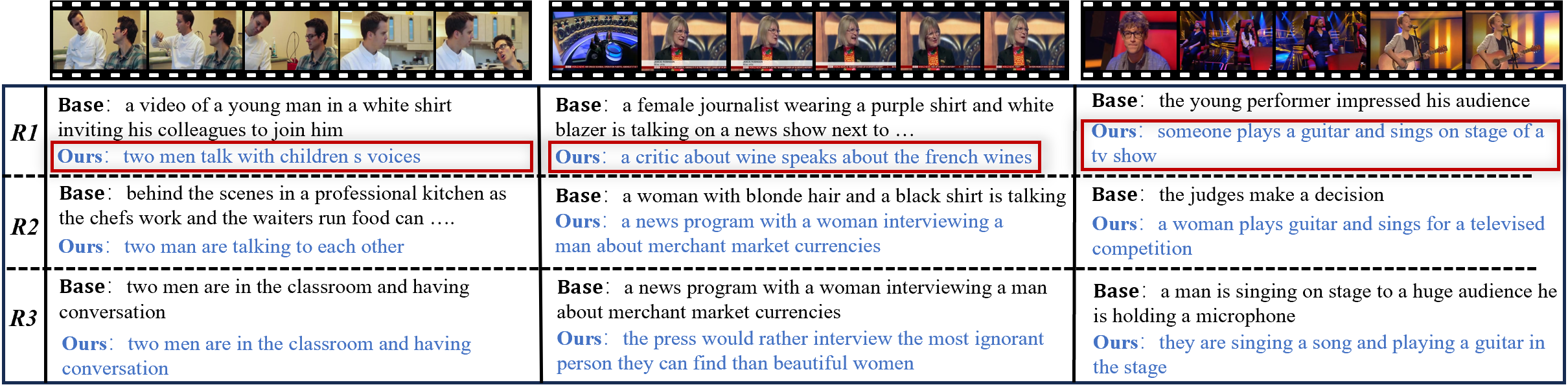}
        \vspace{-0.5em}
	\caption{Examples of video-to-text retrieval results on MSR-VTT dataset.  \textit{(The ground truths are marked in a red box.)}}
	\label{fig.comp_v2t_example}
\end{figure*}

Moreover, we demonstrate several VTR cases to show our results more intuitively.
Fig. \ref{fig.comp_t2v_example} presents the top three videos retrieved when given a specific caption. It is shown that 
our method successfully picks the ground-truth videos in the first place. We appreciate seeing that the second and third hit videos are also highly related to the sentences in terms of content.
In Fig. \ref{fig.comp_v2t_example}, we display the text located at the top of the retrieval list when providing a video. Our results are compared to the ``Base" obtained without any caption enhancement.
% Besides, Figure \ref{fig.comp_v2t_example} compares our video-to-text retrieval results to the ``Base" obtained without any caption enhancement.
The ground-truth texts normally take the first rank when using our method. Compared to the base results, our method preferentially selects samples with semantics and descriptive perspectives similar to the ground truths.
For example, our method successfully recognizes keywords of ``voices", ``critic", and ``guitar" that are missed by the Base model.
This is owing to the ECS module that adaptively learns the cross-modal matching for specific samples, benefiting the recognition of key semantics during video-text retrieval.

\subsection{Plug-in Experiment} 
ExCae is encapsulated and supplemented with existing video-text retrieval methods, whose results are compared to those of the original. As shown in Table \ref{tab:plug-in}, our method successfully increases retrieval results of different video-text retrieval methods. The Top-1 recall accuracy is boosted by \textbf{4.8\%}, \textbf{9.4\%} and \textbf{7.7\%} for DRL, TSR-Net and Clip4Clip, respectively. It is observed that the results of plug-in experiments are similar to those obtained by our method with the same backbone \textit{(refer to Table \ref{tab:msr_comparison})}. This indicates that ExCae has strong adaptability and expandability, which can flexibly combine with different models and constantly upgrade its performance as the technology evolves.

\subsection{Ablation Studies}
Our method expands the expression on the video side; thus, this section respectively evaluates the key components, CSI and ECS modules, under different input data cases. To test the effectiveness of the CSI module, we replace the optimal prompt with the initial one to simulate the ablation of the CSI module.  
The removal of the ECS module is realized by directly encoding all the video-side data as visual embeddings. 
As shown in Table \ref{tab:ablation}, the R@1 is around 50\% when only using the original video as video-side data without adding any components; this result is close to the existing ViT-based methods. The ECS module shows only slight positive effects on the original video.  
Similar results can be found if we merely change the video-side data from the original videos to video-derived captions. An obvious performance boost occurs when the original videos and video-derived captions are input in combination, validating the indispensability of these two data. 
In addition, we find that both CSI and ECS modules facilitate cross-modal retrieval. Despite in the absence of the other one, our method achieves over \textbf{6\%} gain in Top-1 recall accuracy. The effectiveness of the CSI and ECS modules proves the superiority of captions' auto-enhancement and multi-perspective cross-modal matching. The best results are obtained by our full method, where both the CSI and ECS modules play mutually reinforcing roles in promoting video-text understanding.

\subsection{Analytical Experiments}

\begin{table}[t]
    \centering
     \caption{Results of plug-in experiments on MSR-VTT dataset.}
     \vspace{-0.8em}
   \begin{tabular}{l|p{1.5em}<{\centering}p{1.5em}<{\centering}p{2.2em}<{\centering}|p{1.5em}<{\centering}p{1.5em}<{\centering}p{2.2em}<{\centering}}
    % \begin{tabular}{l|ccc|ccc}
        \toprule
        \multirow{2}{*}{Method}  &  \multicolumn{3}{c|}{text-to-video}   & \multicolumn{3}{c}{video-to-text}   \\
        \cline{2-7}
           & R@1 &  R@5&  R@10 & R@1 &  R@5 &  R@10      \\

        \midrule
        DRL \cite{wang2022disentangled}  & 49.4 & 76.4 & 84.2 & 47.0 & 77.1 & 84.4 \\
        ~~~~~~~~ ~ + Ours & 54.2 & 81.4 & 90.2 & 54.4 & 81.1 & 89.2 \\
         
         \midrule  
        TS2-Net \cite{liu2022ts2}  & 47.8 & 76.8 & 85.2 & 47.8 & 76.0 & 84.6 \\
        ~~~~~~~~ ~ + Ours & 57.2 & 86.2 & 92.6 & 56.4 & 85.0 & 91.3  \\
        
        \midrule  
        Clip4Clip \cite{luo2022clip4clip}  & 46.4  & 72.1 &  82.0 &  45.4 &  73.4  & 82.4 \\
        ~~~~~~~~ ~ + Ours & 54.1 & 82.1 & 89.5 & 51.2 & 80.6 & 88.8 \\

    % \hline ExCae
         \bottomrule
    \end{tabular}
    
    \label{tab:plug-in}
\end{table}

\begin{figure}[t]
\centering
\includegraphics[width=0.98\columnwidth]{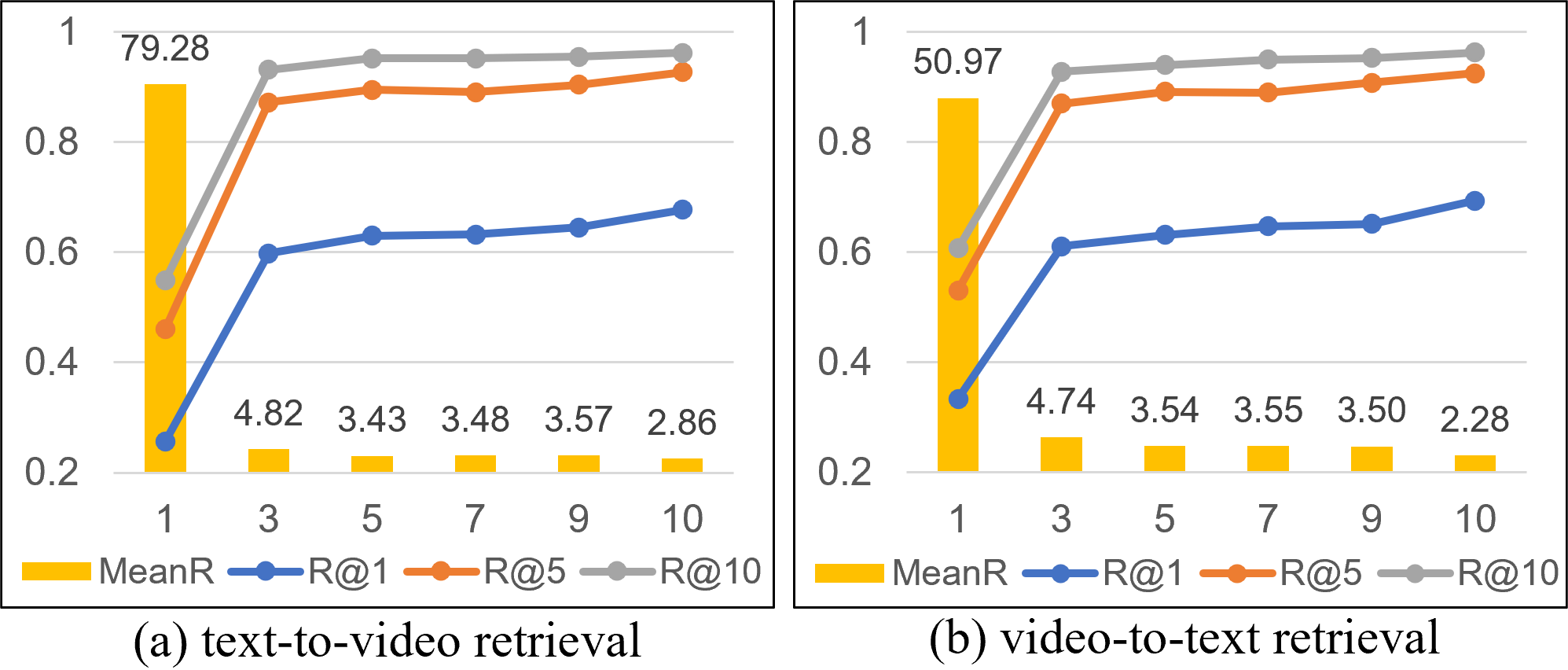} 
\caption{Sensitivity to the number of video-derived captions}
\label{fig:analytical_cn}
\end{figure}

\paragraph{\textit{Effect of video-derived caption number ($N_c$)}}
Fig. \ref{fig:analytical_cn} plots the video-text retrieval results when raising $N_c$ from 1 to 10. The performance is really poor if only one caption is derived, with a Top-1 recall accuracy of lower than 30\% on text-to-video retrieval. This result is inferior to any existing methods, reflecting that insufficient caption enhancement would even damage cross-modal learning. The reason is that all generated captions are forcibly activated to match the text if too few caption samples are input to the ECS module. Consequently, the original text may align with a description from a completely different perspective.
Results are much better when increasing the number of generated captions to 3, achieving good performance close to the existing pre-training models. The performance further improves with the increase of the caption number, and the upward trend becomes stable when $N_c$ is greater than 7. This experiment suggests an approximate range of $[7,10]$ for the selection of video-derived caption numbers.

\begin{figure}[t]
\centering
\includegraphics[width=0.98\columnwidth]{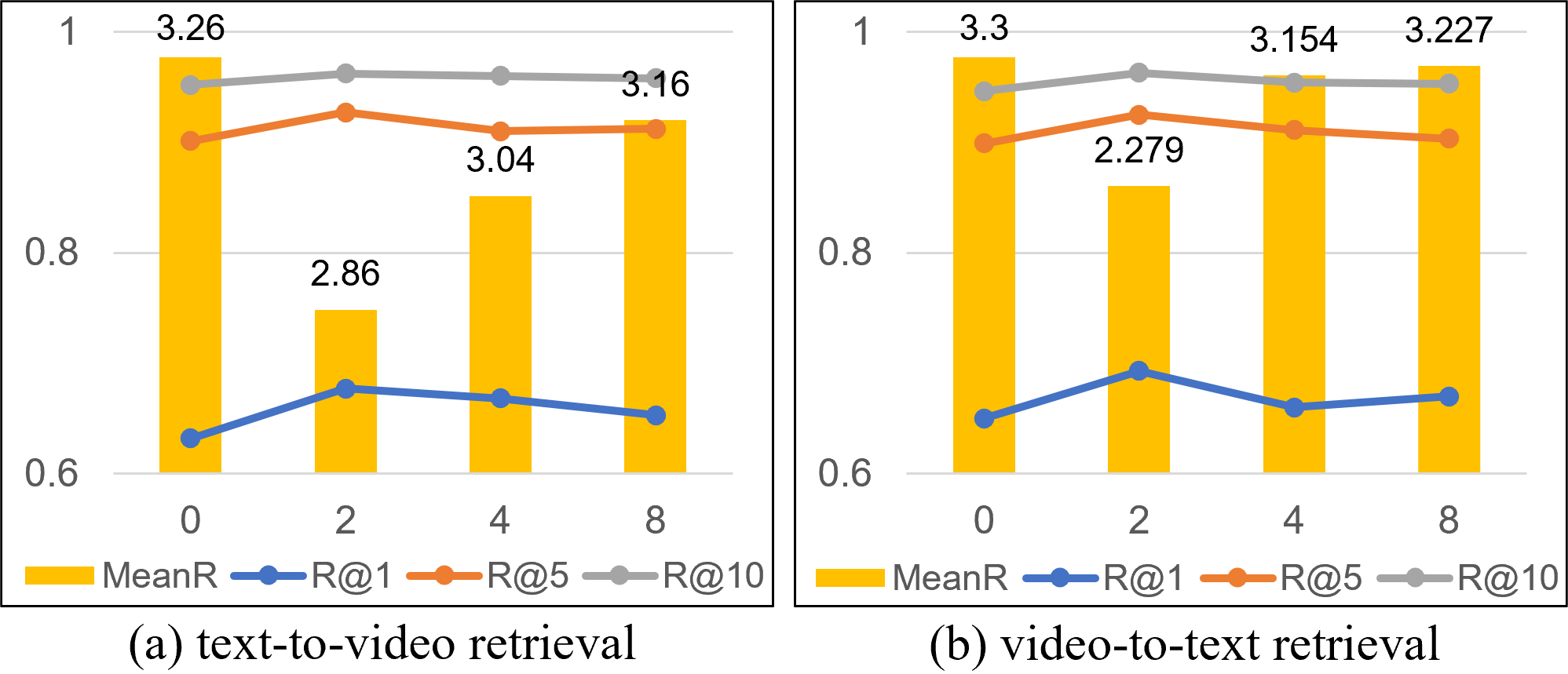} 
\caption{Sensitivity to the number of activated experts}
\label{fig:analytical_en}
\end{figure}

% \paragraph{\textit{b) Effect of caption number}}
\paragraph{\textit{Effect of activated expert number ($N_e$)}}
Video-text retrieval results with different $N_e$ values are displayed in Fig. \ref{fig:analytical_en}. We also show the results obtained without the ECS module, labeled as 0 activated expert, as a reference. The retrieval results gradually improve as $N_e$ increases from 0 to 2. 
However, no further improvement in performance is achieved with the subsequent increase of activated experts. This may be because ECS would automatically favor partial experts and suppress the attention to less relevant expressions, although the majority of experts are activated.

\begin{table*}[t]
	\centering
     \caption{Ablation studies of critical components}
 \small
	\begin{tabular}{c|cc||c|c|c|c|c|c|c|c}
		\toprule
		 &\multicolumn{2}{c||}{ Component}  &  \multicolumn{4}{c|}{text-to-video}   & \multicolumn{4}{c}{video-to-text} \\
   \cline{2-11}
      Video-side Data   &CSI &ECS & R@1$\uparrow$ &  R@5$\uparrow$ &  R@10$\uparrow$ & MeanR$\downarrow$ & R@1$\uparrow$ &  R@5$\uparrow$ &  R@10$\uparrow$    & MeanR$\downarrow$  \\
		\midrule
	 \multirow{2}{*}{Video} & & & 49.9 &	78.1 &	\underline{88.1} &	\underline{7.9}	& 49.6 &\underline{78.8}&	\underline{87.3}	& \underline{7.3} \\
    && \ding{51}&   \underline{52.9}	&\underline{78.5}	& 86.7&	10.1	&\underline{51.3}&	78.5&	87.1	&10.3 \\
  \midrule
	&&& 51.9	& 79.7	&88.3	&9.1	&51.4&	79.8	&87.1&	10.9  \\
   \multirow{2}{*}{ Video-derived Caption}  &\ding{51} && 53.1 &	78.6	&86.7&	10.2	&51.0	&78.6	&87.4&	10.6   \\
	 	&&\ding{51} &   52.3 &	78.2	& 87.9 & 9.0	 & 51.1 &	80.6 &	87.3	& 9.0 \\
		&\ding{51}&\ding{51}&  \underline{54.3}	& \underline{80.3} &	\underline{90.0}	&\underline{6.7}	&\underline{51.8}&	\underline{80.9}	&\underline{88.5}&	\underline{6.8} \\
		\midrule
        & & &   56.2 &	83.7 &	90.9	 &7.2  &	56.9 &	84.8 &	91.4 &	6.6      \\
        Video+ &\ding{51} & &  63.2 &	90.1	&95.2 &	3.3 &	65.0 &	89.9 &	94.6 &	3.3 \\
		\multirow{2}{*}{ Video-derived Caption} 	&&\ding{51} &63.3 &	89.1 &	95.0	 &4.3 &	62.9 &	89.1	 &94.2 &	4.2\\
		
		&\ding{51}&\ding{51}& \underline{67.7} & 	\underline{92.7}	 & \underline{96.2}	 & \underline{2.9}	 & \underline{69.3}	 & \underline{92.5}	 & \underline{96.3}	 & \underline{2.3}\\
		\bottomrule
	\end{tabular}

	\label{tab:ablation}
\end{table*}

\section{Discussion}
\subsection{Superiority}
\begin{figure}[t]
\centering
\includegraphics[width=0.98\columnwidth]{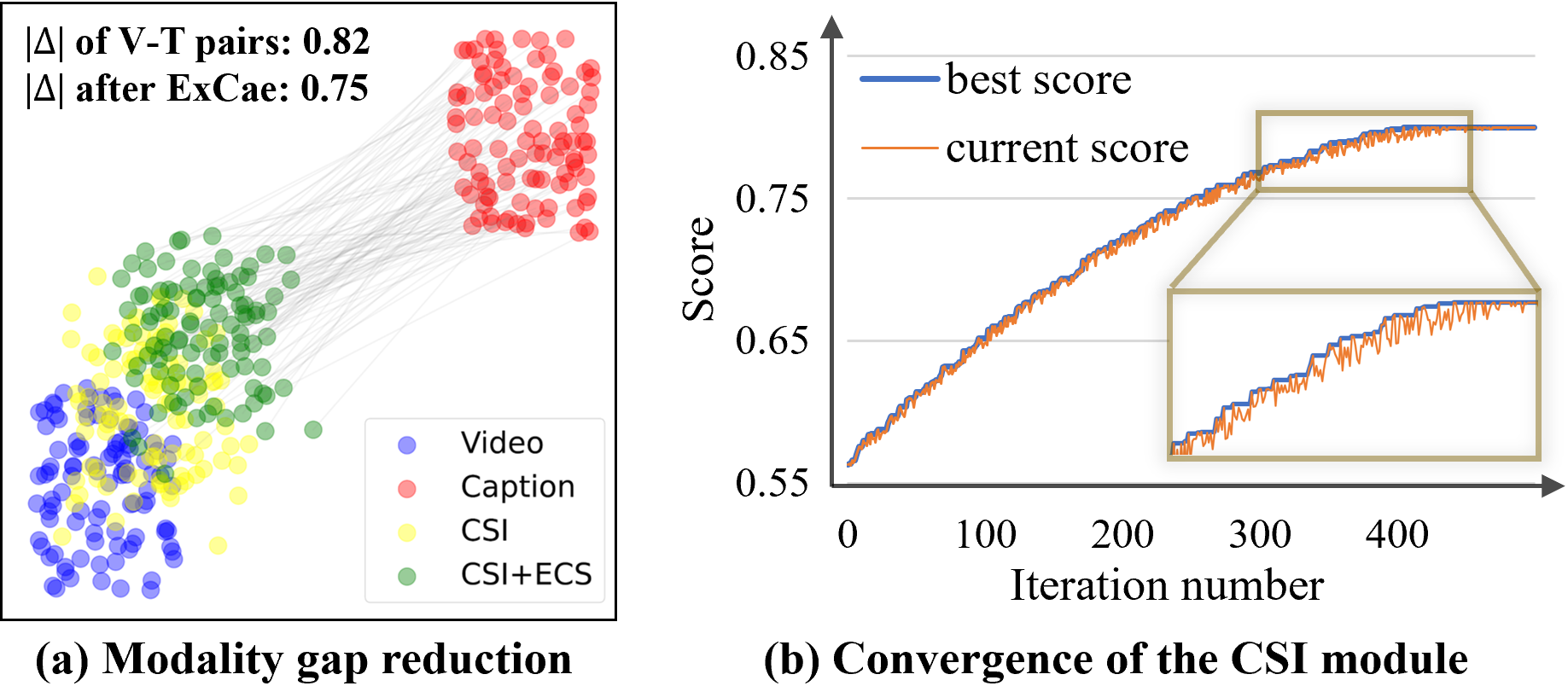} 
\caption{Modality gap and Convergence on MSR-VTT}
\label{fig:convergence}
\end{figure}

ExCae attains remarkable video-text retrieval performance primarily due to its efficient narrowing of the video-text modality gap. As shown in Fig. \ref{fig:convergence} (a), the representations of video-text pairs are drawn closer to each other through self-learning diffusional data augmentation (CSI) and self-adaptive cross-modal matching  (ECS).
The modality gap $||\Delta||$ \cite{liang2022mind} is reduced from 0.82 to 0.75, from 0.79 to 0.67, and from 0.84 to 0.74 on MSR-VTT, MSVD and DiDeMo, respectively.  
Moreover, ExCae has advantages in objectivity, generalization and self-adaptation.
1) ExCae is a \underline{\textit{\textbf{data-driven method}}}, in which the query prompts and caption selection are learned according to the training data. No additional human work \textit{(despite initialization)}, such as prompt design and lexicon establishment, is involved in the whole learning process. Thus, ExCae effectively avoids the adverse effects of empiricism.
2) As the \textit{Captioner} is asked to generate multi-angle captions using a diversity constraint, data with more expression perspectives are provided for model learning. Therefore, ExCae inherits the strength of \underline{\textit{\textbf{generalization improvement}}} from data augmentation.
3) ExCae also has \underline{\textit{\textbf{personalized ability}}} owing to the parametric learning of the ECS module. The router in the ECS module would adaptively select the best expressions for cross-modal retrieval, thus maintaining the robustness of ExCae in different applications.

\subsection{Convergence}
Since the generated captions are changeless once the optimal prompt is determined, the cross-modal post-training of our method owns the same convergence as existing ViT-based methods. 
Regarding learning the CSI module, the LLM would optimize query prompts in the given direction. With the increase of iteration, the semantic changes in query prompts become smaller, and thus the caption evaluation scores tend to be stable \textit{(See Fig. \ref{fig:convergence} (b))}.
According to expertise from experiments, the CSI module tends to converge in 400 iterations.

\subsection{Storage and Efficiency}
Although a prompt optimization process is additionally involved, we believe that the storage and computation cost of ExCae is appreciable, especially compared to the pre-training methods. 
With regard to storage, no extra data or related lexicon needs to be stored. We only need to occupy a very small amount of storage space to record several generated captions in each epoch, which greatly reduces the storage cost. For the computation cost, our method is more efficient than the multi-modal foundation models without a pre-training process. In comparison to the ViT-based methods, the only additional computation is from the training of the CSI module. However, prompt optimization is a one-off training process, and we have also found that the learned prompt has generalization ability across different data sets. Therefore, the efficiency of our approach is comparable to existing VTR methods.

\section{Conclusion}
% This paper proposes strengthening cross-modal understanding by enhancing video expressions in a self-learning manner, which effectively reduces the modality gap without additional workload on data acquisition.
% We innovatively develop a CSI module that gradually refines query prompts to generate high-quality video captions with rich and diverse expressions. The 

% facilitating the while eliminating human empiricism on prompt design.
% % avoids extra workload on data acquisition and prompt design. 
% % A CSI module that gradually refines query prompts is designed to generate high-quality captions with rich and diverse expressions, so as to expand the receptive field for cross-modal learning.
% In addition, multiple learnable experts are introduced as a filter to automatically select the appropriate perspectives of expressions for video-text matching. Our method is non-empirical and self-adaptive in comparison with current text rewriting methods. 
% Experimental results demonstrate that our method achieves SOTA performance on benchmark datasets such as MSR-VTT, MSVD, and DiDemo. Looking ahead, we aspire to explore the variants adaptive to various cross-modal scenarios, thereby aiding a multitude of video-text-related tasks.

This paper proposes strengthening cross-modal understanding by enhancing video expressions rather than text rewriting, which effectively reduces the modality gap without additional workload on data acquisition.
We innovatively developed a CSI module that automatically refines query prompts to generate high-quality video captions with rich and diverse expressions.   The visual domain is expanded to meet the expression modes and semantic information of texts, so as to facilitate cross-modal understanding.
In addition, multiple learnable experts are introduced as a filter to automatically select the appropriate perspectives of expressions for video-text matching.    Our method is non-empirical and self-adaptive in comparison with current text rewriting methods.
Experimental results demonstrate that our method achieves SOTA performance on benchmark datasets such as MSR-VTT, MSVD, and DiDemo.    Looking ahead, we aspire to explore the variants adaptive to various cross-modal scenarios, thereby aiding a multitude of video-text-related tasks.

\section*{Acknowledgments}
This work was supported in part by the National Natural Science Foundation of China (NSFC) under Grant 62472105, in part by the Natural Science Foundation of Guangdong Province under Grants 2024A1515010186 and 2025A1515011385, as well as the Regional Joint Fund Project of the Basic and Applied Basic Research Foundation of Guangdong Province under Grant 2022A1515140096.

\bibliographystyle{IEEEtran}
\bibliography{main}

\end{document}

\maketitle

\subsection*{A. Prompts and examples of video-derived captions}
%We designed an algorithm that leverages GPT-4 to automate the enhancement of prompts for video caption generation. 

Table \ref{tab:prompt} lists the refined prompts obtained by the \textit{Prompt Engineer} in the CSI module. Initially, we start with a basic prompt aimed at producing captions from various perspectives, \textit{i.e.}, \textit{``Generate 10 captions from different perspectives about this video."} With iterations of CSI module, the prompt asking the \textit{Captioner} would include more precise and relevant terms, such as actions, behaviors, and detailed visual elements, as shown in the last row of Table \ref{tab:prompt}. More intuitively, we show examples of generated captions in Table \ref{tab:captions}.

\begin{table}[h]
    \centering
    \begin{tabular}{@{}c|m{6.8cm}@{}}
        \hline
        \textbf{\textit{Iter.}} & \textbf{Prompt} \\
        \hline
        0 & Generate 10 captions from different perspectives about this video. \\
        \hline
        1 & Generate 10 unique captions for the provided video frame, each from a different perspective. \\
        \hline
        2 & ...\\
        \hline
        last/2 & Generate 10 unique captions for the provided video frame, each from a different perspective. Ensure that each caption captures key elements such as actions, emotions, and visual details. The captions should be concise yet descriptive, providing a clear understanding of what is happening in the frame. \\
        \hline
        ... & ... \\
        \hline
         last & Generate 10 unique captions from different perspectives for the provided video frames. The goal is to create captions that accurately reflect the content and context of the video frames, capturing key elements such as actions, emotions, and visual details. Each caption should be concise yet descriptive, providing a clear understanding of what is happening in the frame. The target captions should indicate that this is a car review video. Example target caption:... \\
        \hline
    \end{tabular}
    \caption{Prompt of each iteration in the CSI module}
    \label{tab:prompt}
\end{table}

\subsection*{B. Pseudo code of the CSI module}
Pseudo code of the CSI module is shown in Algorithm \ref{alg:CSI}. Take the initial prompt \textit{(init)} and video-text pairs (data) as inputs, CSI iteratively update the best prompt by comparing current score to the best one until there is no improvement. 

\begin{algorithm}[h]
\caption{Caption Self-improvement (CSI) algorithm}
\label{alg:algorithm}
\begin{algorithmic}[1] 
\REQUIRE init, data
\ENSURE best\_prompt, best\_score
\STATE prompt, best\_prompt $\leftarrow$ init
\STATE best\_score, score $\leftarrow -\infty, -\infty$
\STATE no\_improve, max\_no\_improve $\leftarrow 0, 2$
\WHILE{no\_improve \textless max\_no\_improve}
    \STATE improved $\leftarrow$ \textbf{false}
    \FOR{batch in data}
        \STATE \textit{// Generate captions}
        \STATE g\_captions $\leftarrow$ GenCaps(batch.video, prompt) 
        \STATE \textit{// Current score calculation}
        \STATE score\_sim $\leftarrow$ Sim(batch.caption, g\_captions) 
        \STATE score\_div $\leftarrow$ Div(g\_captions)  
        \STATE score $\leftarrow$ (score\_sim + score\_div) / batch\_size  
        \STATE \textit{// Update for the best score}
        \IF{score \textgreater ~ best\_score}  
            \STATE best\_score $\leftarrow$ score
            \STATE best\_prompt $\leftarrow$ prompt
            \STATE improved $\leftarrow$ \textbf{true}
        \ENDIF
        \STATE prompts $\leftarrow$ RePrompt(batch.video, batch.caption, best\_prompt)
        \STATE prompt $\leftarrow$ SumPrompts(prompts)
    \ENDFOR
    \STATE no\_improve $\leftarrow$ 0 \textbf{if} improved \textbf{else} no\_improve + 1
\ENDWHILE

\RETURN best\_prompt, best\_score

\end{algorithmic}
\label{alg:CSI}
\end{algorithm}

\begin{table*}[ht!]
	\centering
	\begin{tabular}{c|l|p{12cm}}
	\hline
  Dataset &  ID & Video-derived captions\\
    \hline
  \multirow{10}{*}{} & \multirow{10}{*}{} & 1. Women conversing in an office, one asking the other to hand over a cigarette pack. \\
   & & 2. A woman with curly hair smiles as she responds to a question in the office. \\
   & & 3. A serious woman with black hair asks for confirmation about giving her cigarette pack. \\
   & & 4. The black-haired woman shows a shocked or confused expression in the office. \\
   MSR-VTT&video6480 & 5. The black-haired woman driving a car, looking stern, while surrounded by motorcycles in traffic. \\
   & & 6. A close-up of the woman driving, looking alarmed as motorcycles pass by her car. \\
   & & 7. A silhouette of the woman peering through a doorway into a dark room. \\
   & & 8. Walking down a dimly lit street, a man tells the woman that introducing him will resolve a conflict. \\
   & & 9. The woman and the man continue their conversation about introductions while walking at night. \\
   & & 10. The woman gestures emphatically towards the man as they discuss introductions on the nighttime street. \\
        \hline
   \multirow{10}{*}{} & \multirow{10}{*}{} & 1. A man in a blue shirt demonstrates precision, swinging a sword towards a water bottle in a serene outdoor setting. \\
   & & 2. The moment of impact captured as the sword meets the water bottle, sending a splash into the air. \\
   & & 3. Focused and intense, the man prepares for his next move against the stationary water bottle. \\
   & & 4. In a split second, the water bottle begins to shatter under the force of the expertly wielded sword. \\
   MSVD & -8y1Q0rA3n8 & 5. Partially sliced, the water bottle stands, seconds before it succumbs to the powerful swing. \\
   & \_108\_115 & 6. A decisive swing catches the water bottle in mid-air, showcasing the mans skill and control. \\
   & & 7. Captured mid-action, the clear arc of the sword as it slices through the bottle in an outdoor environment. \\
   & & 8. Splashes of water glisten in the sunlight as they are flung into the air from the force of the sword strike. \\
   & & 9. The cut through the water bottle is clean, demonstrating the sharpness of the sword and the precision of the man wielding it. \\
   & & 10. Amidst the greenery and under bright skies, the sword’s blade comes down on the water bottle in a display of martial skill. \\
	\hline
   \multirow{10}{*}{} &  & 1. A busy kitchen counter with a glistening knife and a bucket of coleslaw nearby, ready for meal preparations. \\
   & & 2. Thin slices of succulent brisket fall onto the wooden counter, the chef meticulous with each cut. \\
   & & 3. A gloved hand expertly carves the meat, revealing the juicy pink interior of the brisket. \\
   & 80591230 & 4. Seasoned brisket is carefully sliced, the chef ensuring each piece is perfectly cut for serving. \\
   DiDeMo & @N00\_3471997189 & 5. Chunks of meat and crumbs scatter across the counter as the brisket is dissected with precision. \\
   &\_4394214f13 & 6. The kitchens bustling environment captured in the midst of preparing a delicious brisket meal. \\
   &  & 7. Well-worn knife and coleslaw bucket set the scene for a busy day in the kitchen. \\
   & & 8. Brisket being sliced with finesse, ready to be plated alongside fresh slices of bread. \\
   & & 9. The chefs hands move quickly and efficiently, turning a large chunk of brisket into thin, delectable servings. \\
   & & 10. Using a knife to precisely slice through meat. \\
	\hline
	\end{tabular}
 \caption{Examples of video-derived captions}
	\label{tab:captions}
\end{table*}
\begin{figure*}[h]
	\centering
	\includegraphics[width=0.98\textwidth]{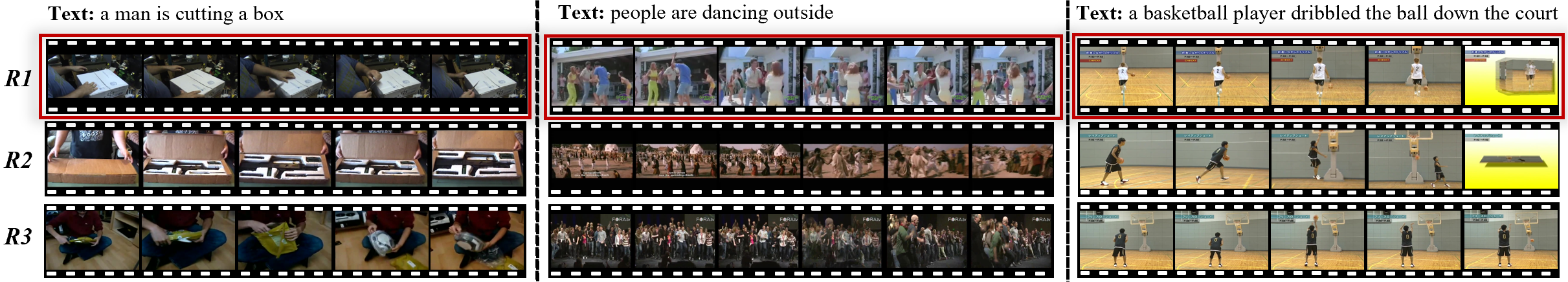}
        % \vspace{-0.5em}
	\caption{Examples of text-to-video retrieval results on MSVD dataset.  \textit{(The ground truths are marked in a red box.)}}
	\label{fig.msvd_comp_t2v_example}
\end{figure*}
\begin{figure*}[h]
	\centering
	\includegraphics[width=0.98\textwidth]{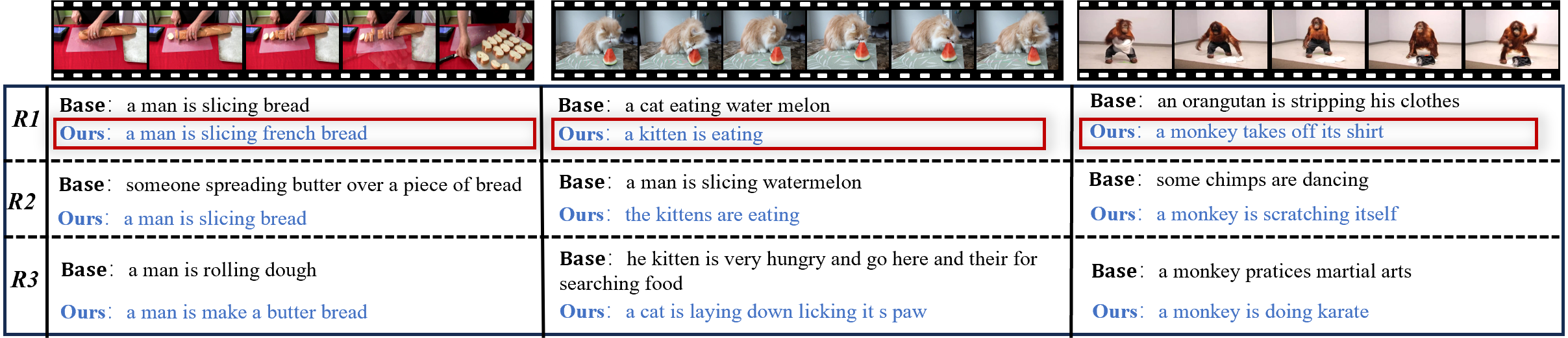}
        % \vspace{-0.5em}
	\caption{Examples of video-to-text retrieval results on MSVD dataset. Base: without caption enhancement; Ours: ExCae;  \textit{(The ground truths are marked in a red box.)}}
	\label{fig.msvd_comp_v2t_example}
\end{figure*}
\begin{figure*}[h]
	\centering
	\includegraphics[width=0.98\textwidth]{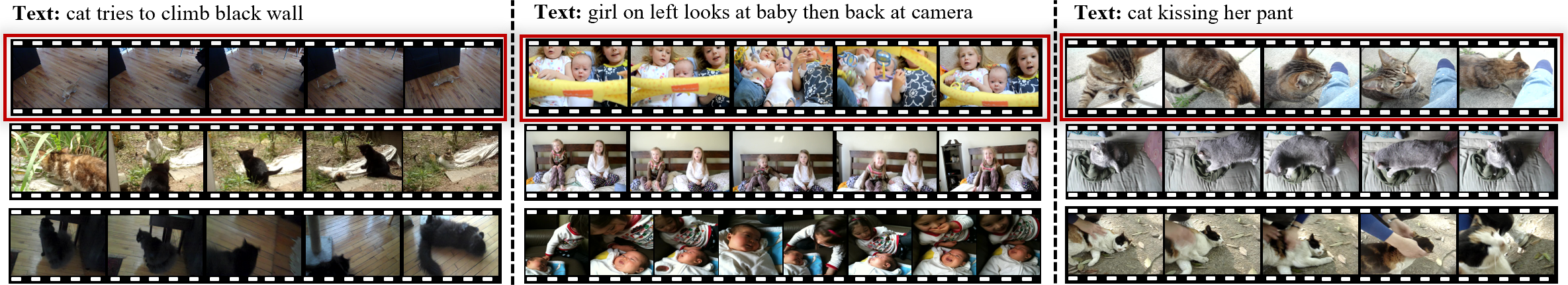}
        % \vspace{-0.5em}
	\caption{Examples of text-to-video retrieval results on DiDeMo dataset.  \textit{(The ground truths are marked in a red box.)}}
	\label{fig.didemo_comp_t2v_example}
\end{figure*}
\begin{figure*}[h]
	\centering
	\includegraphics[width=0.98\textwidth]{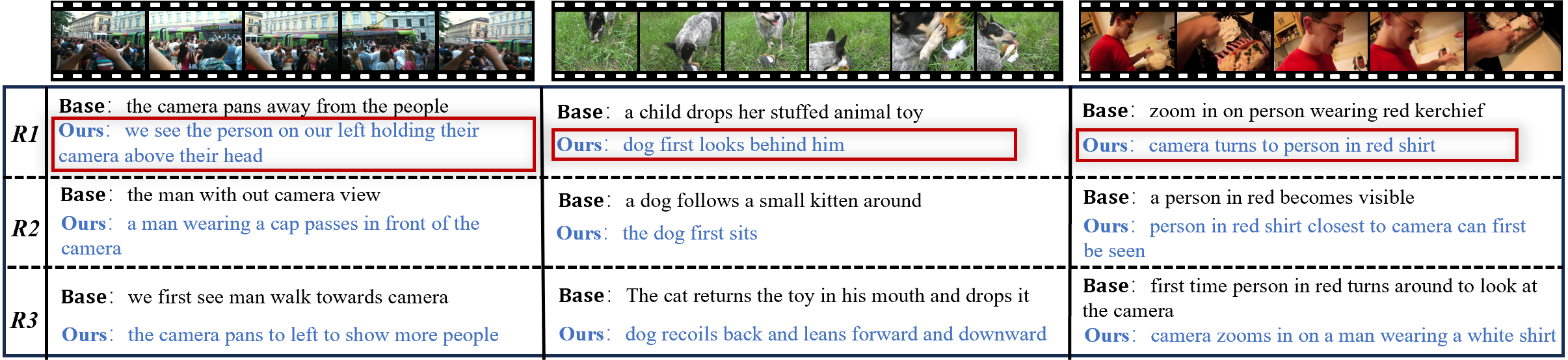}
        % \vspace{-0.5em}
	\caption{Examples of video-to-text retrieval results on DiDeMo dataset. Base: without caption enhancement; Ours: ExCae;  \textit{(The ground truths are marked in a red box.)}}
	\label{fig.didemo_comp_v2t_example}
\end{figure*}
\clearpage

\subsection*{C. Detailed formula of caption scoring}
In the original formula \textit{(Eq. (1) in the manuscript)} for caption scoring, the term $sim(\cdot)$ computes the similarity between generated captions and original text. In specific, $sim(\psi(c^t_{i,k}),\psi(t_i))=\frac{\psi(c^t_{i,k}) \cdot \psi(t_i)}{\|\psi(c^t_{i,k})\| \|\psi(t_i)\|}$.
Moreover, the term \( \text{div}(\psi(c^t_i)) \) calculates the diversity of samples in the caption list, which can be defined as \( \text{div}(S_i) \), where \( S_i \) is the similarity matrix between captions, defined as \( S_i[p,q] = \text{sim}(\psi(c^t_{i,p}), \psi(c^t_{i,q})) \), \( p \neq q \). Then, $div(\psi(c_i^t))$ can be rewritren as:

\begin{equation}
div(\psi(c_i^t))= \frac{1}{K(K-1)} \sum_{p \neq q} (1 - S_i[p,q])
\end{equation}where $K$ denotes the caption number in the caption list.This definition ensures that higher diversity is achieved when the similarity between different captions is lower, thus effectively promoting multi-angle descriptions.

\subsection*{D. Examples of video-text retrieval results}
Examples of video-text retrieval results on MSVD and DiDeMo datasets are displayed in Figures \ref{fig.msvd_comp_t2v_example} to  \ref{fig.didemo_comp_v2t_example}. 
These examples show results consistent with observations obtained in the MSR-VTT dataset.

\subsection*{E. Instructions for Special cases}
\enlargethispage{\baselineskip}
\enlargethispage{\baselineskip}
\enlargethispage{\baselineskip}
\enlargethispage{\baselineskip}
\enlargethispage{\baselineskip}
\enlargethispage{\baselineskip}
\enlargethispage{\baselineskip}
\enlargethispage{\baselineskip}
When using GPT-4o to generate multi-angle captions for videos, it may trigger GPT-4o's safety protocols if the video contains minor adult content, causing failure in caption generation. To address these specific cases, we consider an offline fallback operation and deploy an offline multimodal large model (InternVL2-8B) to substitute GPT-4o for handling these few video issues.

\subsection*{F. Source codes}
The source codes of this paper is summarized in code.zip, zipped in the supplemental materials.

\clearpage